\documentclass[runningheads]{llncs}
\usepackage{graphicx}

\usepackage{tikz}
\usepackage{comment} 
\usepackage{amsmath,amssymb} 
\usepackage{color}

\usepackage{multirow}
\usepackage{graphicx,xparse,booktabs}

\ExplSyntaxOn
\NewDocumentEnvironment{places}{mm}
 {
  \setlength{\tabcolsep}{-0pt} 
  \dim_set:Nn \l_places_width_dim
   {
    (#1-\ht\strutbox-\dp\strutbox-2pt)/(#2)
   }
  \begin{tabular}{l *{#2}{c}}
 }
 {
  \end{tabular}
 }

\NewDocumentCommand{\place}{mm}
 {
  \seq_set_from_clist:Nn \l_places_images_in_seq { #2 }
  \seq_set_map:NNn \l_places_images_out_seq \l_places_images_in_seq { \places_set_image:n {##1} }
  \seq_put_left:Nn \l_places_images_out_seq
   {
   }
  \seq_use:Nn \l_places_images_out_seq { & }
  {
  \begin{tabular}{c}\rotatebox[origin=c]{90}{\strut#1}\end{tabular}
  }
  \vspace{-0.05cm}
  \\ 
 }
 
\dim_new:N \l_places_width_dim
\seq_new:N \l_places_images_in_seq
\seq_new:N \l_places_images_out_seq

\cs_new_protected:Nn \places_set_image:n
 {
  \makebox[\l_places_width_dim]
   {
    \begin{tabular}{c}
    \includegraphics[
      width=0.9\l_places_width_dim,
      height=\l_places_width_dim
    ]{#1}
    \end{tabular}
   }
 }
\ExplSyntaxOff

\begin{document}
\pagestyle{headings}
\mainmatter
\def\ECCVSubNumber{100}  

\title{Object Retrieval and Localization \\ in Large Art Collections using \\ Deep Multi-Style Feature Fusion and \\\ Iterative Voting} 

\titlerunning{Object Retrieval and Localization in Large Art Collections}
%
\author{Nikolai Ufer \and Sabine Lang \and Bj\"orn Ommer}
%
\authorrunning{N. Ufer et al.}
%
\institute{
Heidelberg University, HCI / IWR, Germany \\
\email{nikolai.ufer@iwr.uni-heidelberg.de, ommer@uni-heidelberg.de} 
}
\maketitle
\begin{abstract}
The search for specific objects or motifs is essential to art history as both assist in decoding the meaning of artworks. Digitization has produced large art collections, but manual methods prove to be insufficient to analyze them. In the following, we introduce an algorithm that allows users to search for image regions containing specific motifs or objects and find similar regions in an extensive dataset, helping art historians to analyze large digitized art collections. 
Computer vision has presented efficient methods for visual instance retrieval across photographs. However, applied to art collections, they reveal severe deficiencies because of diverse motifs and massive domain shifts induced by differences in techniques, materials, and styles. In this paper, we present a multi-style feature fusion approach that successfully reduces the domain gap and improves retrieval results without labelled data or curated image collections. Our region-based voting with GPU-accelerated approximate nearest-neighbour search [29] allows us to find and localize even small motifs within an extensive dataset in a few seconds. We obtain state-of-the-art results on the Brueghel dataset [2,\,53] and demonstrate its generalization to inhomogeneous collections with a large number of distractors.
\end{abstract}

\section{Introduction}
For art history, it is crucial to analyze the relationship between artworks to understand individual works, their reception process, and to find connections between them and the artists \cite{johnson2012memory,hristova2016images,impett2016pose}. Hereby, the investigation of motifs and objects across different images is of particular importance since it allows more detailed analyses and is essential for iconographic questions. 
Digitization has produced large image corpora \cite{wikiart,artuk,wga,strezoski2017omniart}, but manual methods are inadequate to analyze them since this would take days or even months.
Computer-assisted approaches can dramatically accelerate and simplify this work. 
However, most of them consist of a simple text search through metadata and are not sufficient since text cannot capture the visual variety, and labels are often either missing, incomplete, or not standardized. Therefore, there is a need for efficient algorithms, capable of searching through visual art collections not only based on textual metadata but also directly through visual queries. In this paper, we present a novel search algorithm that allows users to select image regions containing specific motifs or objects and find similar regions in an extensive image dataset.

While computer vision successfully developed deep learning-based approaches for visual instance retrieval in photographs \cite{radenovic2018revisiting,radenovic2018fine}, artworks present new challenges. 
This includes a domain shift from real photos to artworks, unique and unknown search motifs, and a large variation within art collections due to different digitization processes, artistic media, and styles. This strongly highlights the need for specifically tailored algorithms for the arts \cite{seguin2017tracking,shen2019discovering}. 
Solving visual instance retrieval across artworks is difficult and requires local descriptors, which are both highly discriminative to find matching regions and invariant regarding typical variations in art collections. Learning such descriptors in a supervised fashion is extremely time-consuming and requires annotating thousands of corresponding images \cite{seguin2016visual}.
Besides, the learned descriptors improve retrieval results only for very similar datasets.
Alternative approaches based on self-supervision \cite{shen2019discovering} show promising results. However, they are not stable against images without any repetitions in the dataset, and are very slow in large-scale scenarios since they require a pairwise comparison between all images. 
We circumvent these issues and present a new multi-style feature fusion, where we utilize generic pre-trained features and current style transfer models to improve their style invariance without any additional supervision. Given a dataset, we stylize all images according to a set of fixed style templates, and by mixing their feature representations, we project them into the same averaged style domain. This massively reduces the domain gap across artworks and improves overall retrieval results. 

\textbf{Contributions.} 
Our main contributions are threefold. (1) We present an unsupervised multi-style feature fusion, which successfully reduces the domain gap and improves overall retrieval results in art collections. (2) The introduced iterative voting in combination with GPU-accelerated nearest-neighbour search \cite{faiss17} enables us to localize and find small regions in large datasets within a few seconds. (3) We demonstrate that the proposed method significantly outperforms current methods in terms of retrieval time and accuracy for object retrieval and localization in art collections.

\section{Related work}
In the following, we present the most relevant research related to our work and put our contributions in context.

\textbf{Computer vision in the arts.}
For quite some time, there has been a mutual exchange between computer vision and the arts. This exchange ranges from the analysis of artworks using computer vision \cite{arora2012towards,crowley2014,seguin2016visual,gonthier2018weakly} to the development of new methods in collaboration with the art community \cite{schlecht2011detecting,ufer2012wehrli,takami2014approach,shen2019large} to generative algorithms that transfer normal photos into artworks \cite{hertzmann2001image,gatys2016image,sanakoyeu2018style} or direct attempts to create art \cite{elgammal2017can,hertzmann2018can}. 
Concerning the analysis of artworks, collaborations are very natural because computer vision and art history are both concerned with the visuality and ask similar questions. In this context, researchers have transformed successful object detection and classification methods for photos to artworks \cite{schlecht2011detecting,crowley2014,karayev2013recognizing,spratt2014computational}. Some of these works detect gestures, people, or iconographic elements in paintings \cite{schlecht2011detecting,ginosar2014detecting,westlake2016detecting,yin2016object,gonthier2018weakly,jenicek2019linking}, recognize object categories occurring in natural images \cite{crowley2014,crowley2016,wilber2017bam}, classify paintings in terms of their style, genre, material, or artist \cite{karayev2013recognizing,mensink2014rijksmuseum,saleh2015large,saleh2016toward,tan2016ceci,wilber2017bam,strezoski2017omniart}, or investigate the aesthetics of paintings \cite{spratt2014computational,amirshahi2014jenaesthetics,denzler2016convolutional}.
Some approaches directly try to find visual relationships within art collections automatically \cite{seguin2016visuallink,shen2019discovering,garcia2020contextnet} to relieve as much work as possible from art historians.
However, these are very time-consuming since they compare all possible image pairs within the dataset.
 Therefore, they are limited to small collections, which could also be manually analyzed. 
In this work, we focus on large-scale instance retrieval and localization in the arts. We are convinced, with an efficient search system, art historians can find relevant visual links through several searches faster and more targeted compared to fully automated approaches.

\textbf{Visual instance retrieval.}
Visual instance retrieval deals with the task of identifying matching regions in other images within a dataset, given a query region or image. This is a well-established research field in computer vision with successful classical \cite{jegou2008hamming,crowley2014}, as well as deep learning-based \cite{tolias2015particular,yi2016lift,noh2017large} approaches.  
Early methods were based on classical feature point descriptors like SIFT \cite{lowe2004distinctive} combined with a Bag-of-Words approach \cite{jegou2008hamming,crowley2014}. Over the years, numerous improvements have been made for different parts of this approach \cite{shen2012object}. 
More recently, Convolutional Neural Networks (CNN) showed remarkable results in many areas of computer vision, including visual instance retrieval \cite{collins2019deep}.  
However, the primary retrieval research was always focused on photographs, either of the same place \cite{philbin2007object,philbin2008lost} or the same object \cite{nister2006scalable}. If researchers dealt with art databases, then often only on an image level without allowing to search for regions \cite{shrivastava2011data,takami2014approach,picard2015challenges,crowley2015face,mao2017deepart}, which is a central requirement to find similar motifs and objects in art collections. 
Just recently, Shen et al. \cite{shen2019discovering} introduced the first benchmark with annotations for finding and localizing objects and motifs in artworks on a region level, which is also the primary dataset we use for our evaluation. 
Most closely related to our work is Shen et al. \cite{shen2019discovering} and Seguin et al. \cite{seguin2017tracking}, which also deal with instance retrieval and finding visual relationships in art collections. In contrast to our work, Seguin et al. \cite{seguin2017tracking} use off-the-shelf CNNs that have been fine-tuned in a supervised fashion and thus does not apply to other datasets.  The approach of Shen et al. \cite{shen2019discovering} learns dataset-dependent features in a self-supervised manner by mining correspondences between image pairs, similar to \cite{ufer2017deep,ufer2019weakly}. However, their approach does not generalize to inhomogeneous collections with many distractors, and their retrieval system is intractable in large-scale scenarios.  In contrast, our method improves retrieval results by successfully reducing the domain gap without labelled data or curated image collections and enables us to find even small motifs within an extensive dataset in a few seconds.

\section*{Method}
The following requirements on the image representation and the retrieval system are essential. 
First, it should be possible to search for any motif in a diverse art collection. This poses particular challenges to the underlying feature descriptors since off-the-shelf models are trained on photographs and are not invariant to colours and artistic styles. We address this problem in the first part of this section and introduce a new multi-style feature fusion, which considerably reduces the domain gap. 
Second, it should be possible to search for any image region with an arbitrary size across a large dataset, and the search should take only a few seconds and deliver exact retrieval results. 
These are challenging demands since a single feature descriptor is not capable of capturing multiple objects or motifs in an image accurately, and encoding all regions is not tractable due to memory constraints. We address this problem in the second part of this section. For the query region and all images in the dataset, we extract a moderate number of local descriptors and formulate the search as a voting procedure of local patches within the query region.

\begin{figure}[!t]
\centering
\includegraphics[width=0.98\linewidth]{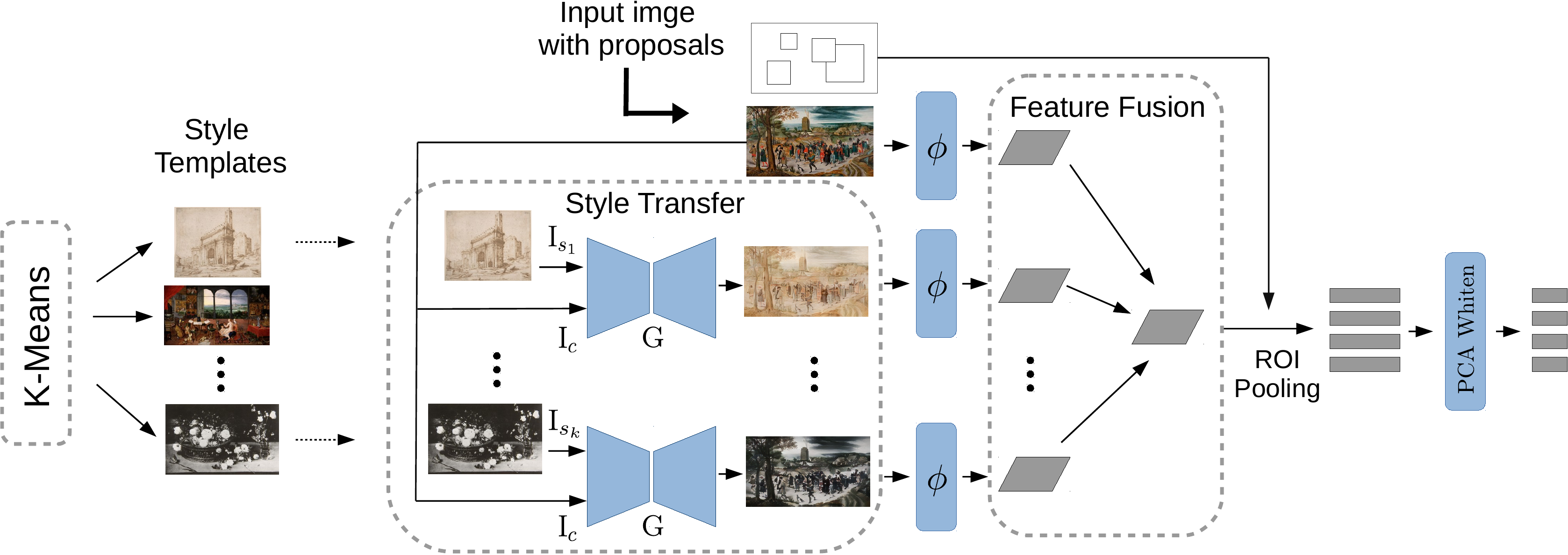}
\caption{
Overview of our multi-style feature fusion. It consists of three main steps: First, we extract image level features of all dataset images, apply K-means and select the cluster centers as style templates. Given an input image and region proposals, the input image is stylized based on the style templates and features are generated using the pre-trained feature extraction and style transfer network. The image features are fused to create a final image representation, and through ROI pooling, we obtain region descriptors for all proposals with a fixed dimension. Finally, we apply principle-component analysis (PCA) and whitening to reduce the feature dimension.
}
\label{fig:msf_overview}
\end{figure}

\subsection*{Multi-style feature fusion}
Our multi-style feature fusion is based on the following hypothesis. The dataset contains images or regions that are similar to each other but are depicted in different styles. Generic pre-trained descriptors are capable of finding similar regions if the style differences are small. Based on this assumption, our main idea is to use current style transfer models to project all images into the same averaged style domain to reduce the domain gap and simplify the retrieval task. Therefore, we stylize each image based on multiple fixed style instances and fuse their extracted features to generate a single robust representation. The approach consists of three main steps, which we describe in the following. See Fig. \ref{fig:msf_overview} for an overview. 

\textbf{Finding style template.} 
To find a diverse set of style templates, we proceed as follows.  We denote the collection of images as $\mathcal{I} = \{I_i \vert 1 \le i \le N\}$ and assume that an ImageNet pre-trained CNN, which maps images into an $m$-dimensional deep feature space, i.e. $\phi \colon \mathcal{X} \to \mathbb{R}^m$, is given. 
We group all images into $k_s$ clusters according to their pairwise distance in the embedding space using K-means. Since the CNN $\phi$ is trained on real photos, it is not invariant with respect to styles, and hence different clusters contain different content depicted in different artistic styles. We select $k_s$ images which are closest to the cluster centers in the feature space and obtain a diverse set, which serves as our style templates $\mathcal{S} = \left\{I_s \vert 1 \le s \le k_s \right\}$. 
This selection is sufficient because our style transfer method is independent of the image content, and only the depicted style is important.
In our experiments, we set $k_s=3$ since this has proven to be a good trade-off between performance and computational cost.

\textbf{Style transfer.} 
Our style transfer module is a CNN, $G \colon \mathcal{X} \times \mathcal{X} \to \mathcal{X}$, which takes a content image $I_c$ and style image $I_s$ as input to synthesize a new image $G(I_c, I_s)$ with the content from the former and style from the latter.  
It is based on the network architecture of Li et al. \cite{li2018learning} and consists of three main parts: an encoder-decoder, a transformation, and a loss module. 
The encoder consists of the first layers of the ImageNet pre-trained VGG-19 model \cite{deng2009imagenet,simonyan2014very} and the decoder comprises of its symmetrical counterpart. 
The transformation module consists of two small CNNs which receive the encoded feature maps of the content and style image as input, and provide a transformation matrix as output, respectively. 
Given a content and style image, the style is transferred by multiplying the encoder's content feature with the two transformation matrices and applying the decoder on the output, which produces the stylized image. 
Since the network is a pure feed-forward convolution neural network, it allows converting an image into an arbitrary style in milliseconds. 

\textbf{Feature fusion.}
Based on the style transfer module $G$ and style templates $\mathcal{S}$, we obtain the multi-style feature representation as follows.
Given an input image $I$ and a set of proposals, we stylize the image with respect to all style templates $\mathcal{S}$, and fuse them by taking their mean, i.e.
\begin{equation}
\phi_{ms}\left(I, \mathcal{S} \right) = \frac{1}{1 + \vert \mathcal{S} \vert} \Bigg( \phi (I) + \sum_{I_s \in \mathcal{S}} \phi\left(G(I, I_{s})\right)  \Bigg).
\end{equation}
Here we also take the original image feature into account since it contains fine-grained information that can be useful for the retrieval task but are lost during the transformation process.
Given the proposals and the new image representation, we apply Precise ROI Pooling \cite{jiang2018acquisition} to obtain local feature descriptors for all proposals with a fixed feature dimension. Finally, we apply principle-component analysis (PCA) and whitening to reduce their feature dimension.

\begin{figure}[!t]
\centering
\includegraphics[width=0.99\linewidth]{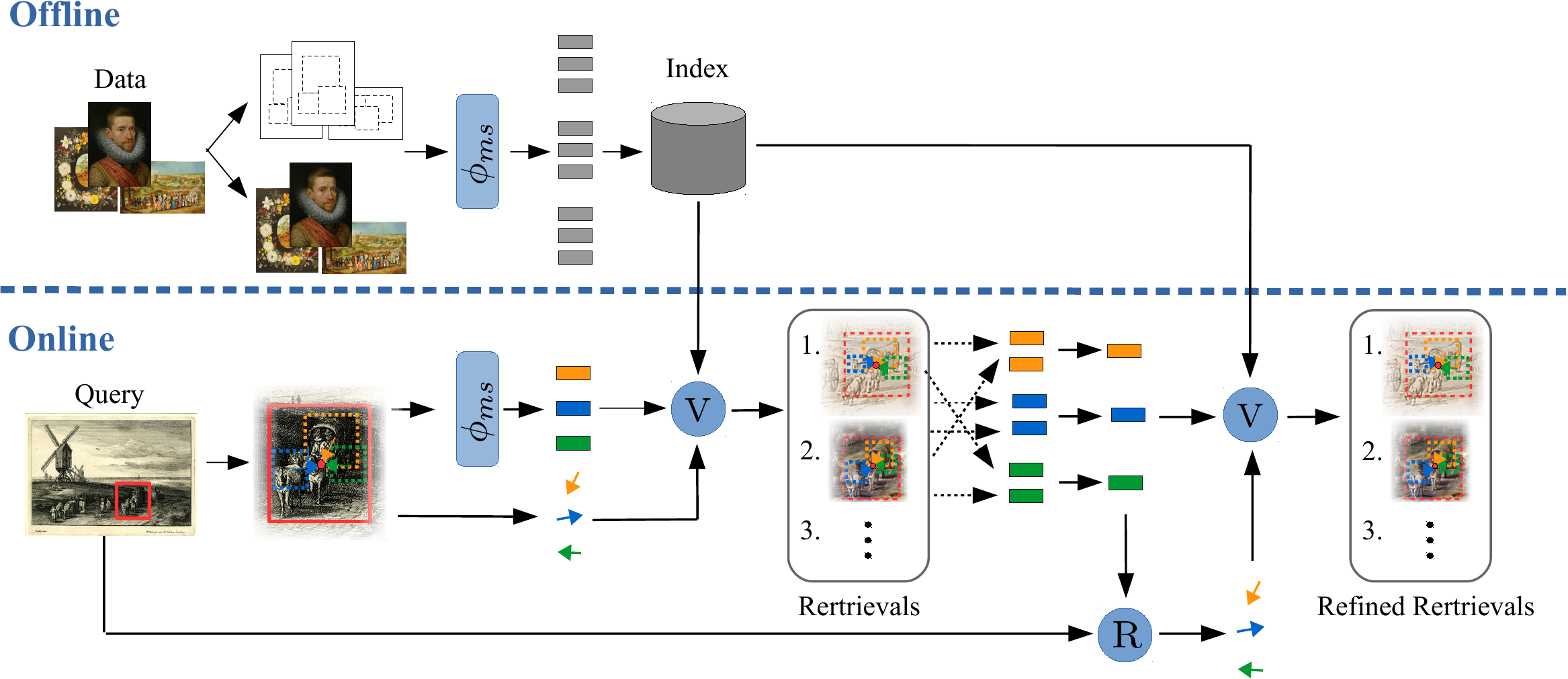}
\caption{
Overview of our retrieval system. The retrieval system consists of an offline and an online stage. During the offline stage, the search index is initialized before the actual search. For each image, discriminative local patches are determined, and their feature descriptors are extracted, compressed, and stored in the search index. During the online stage, most discriminative local patch descriptors within the marked query region are extracted, and their k-nearest neighbours are determined using the search index.
Our voting procedure aggregates these local matches and predicts well-localized retrieval bounding boxes for the whole query. The results are further improved using local query expansion and re-voting. 
}
\label{fig:voting_overview}
\end{figure}

\subsection*{Retrieval system using iterative voting}
To find and localize objects and motifs of arbitrary size in an extensive art collection, we introduce an iterative voting approach based on local patch descriptors. Therefore, we decompose all images into a set of quadratic patches on multiple scales in a sliding-window manner, which are encoded by the multi-style feature extraction network described previously. The actual search consists of finding k-nearest neighbours of local patches within the selected query region across the whole dataset, which are afterwards aggregated to well-localized retrieval bounding boxes. This approach has three main advantages. 
First, the search is performed on a region level and enables searching for small motifs that cannot be found when images are represented with a single feature descriptor. Second, it allows us to predict well-localized retrieval bounding boxes independent of the local patches we use for the image encoding. Third, we combine several search queries in combination with a spatial verification, which provides better search results. 

Our retrieval system consists of an offline and online stage. In the offline stage, the search index is initialized. Therefore, local image descriptors are extracted, compressed, and stored in the index. This step only needs to be done once. In the online stage, most discriminative local feature descriptors within the query region are extracted, and our voting procedure is applied to their k-nearest neighbours to obtain the first search results. Through local query expansion and re-voting, the retrievals results are further improved. In the following, we describe each step in more details. We also provide an overview in Fig. \ref{fig:voting_overview}. 

\textbf{Image encoding using local patch descriptors.}
In our experiments classic region proposal algorithms \cite{manen2013prime,uijlings2013selective,zitnick2014edge} and networks \cite{ren2015faster} showed insufficient results for artworks. 
Our strategy for finding suitable regions to extract local patch descriptors consists of two steps. First, we generate a broad set of local patches by dividing the space into quadratic regions on multiple scales in a sliding window manner. Second, we filter them for selecting the most discriminative as follows. We sample a subset of all local patches and assign them into $k_e$ groups using the $L_2$ distance in the multi-style feature space and K-means. We store the $k_e$ cluster centers, compute the distance of all local patches to these centers, and select those with the highest mean distance. By this, we obtain local patches on different scales that are more discriminative and more suitable for the retrieval task than others. We generate a maximum of 4000 local patches for each image $I$ in this way and extract their multi-style feature descriptors $\mathcal{D}(I)$ as described previously. 
To keep the search index compact, we reduce the number of extracted proposals per image linearly with the dataset size. 

\textbf{Search index.}
We build a search index containing all local patch descriptors of all images $\mathcal{D}$ for fast approximate nearest neighbor search. For the search index, we use the Inverted File Index (IVF) and the Product Quantization (PQ) algorithm from \cite{faiss17}.  They provide a high-speed GPU-parallelized variant that allows to search for multiple queries in an extensive database in seconds. The IVF algorithm \cite{jegou2010product} clusters the feature vectors into groups and calculates their centroids using K-means. Now, given a query vector, the distances to all centroids are determined, and only the feature vectors assigned to the closest centroids are considered for the k-nearest neighbor search, which massively accelerates the search. The actual search uses product quantization. Here the feature vectors are sliced into subvectors, and a codebook for each of these slices is learned. Based on these codebooks, the feature vectors can be stored efficiently using their ids, and a look-up table containing all codebook centroids distances allows a fast approximated nearest neighbor search with a query vector. We calculate the IVF clustering and PQ codebooks for each dataset separately. Therefore, we take all regions from 1000 randomly selected images and train on their multi-style features. 

\textbf{Query reformulation using local patch descriptors.}
In the online stage, the user selects an image and marks a rectangle $q$ as the query region. Besides the selected region itself, we additionally use the most discriminative local patches within the query region for the actual search. We find these local query patches as follows. We extract local patches on multiple scales and select the most discriminative using two criteria. First, we filter all local patches with less than 90 percent overlap with the query region. Second, we apply non-maximum suppression concerning their feature activation, which we obtain by summing over the feature channel and taking the mean within the proposal region. For the selected patches, we extract their features and store their voting vectors. This set of local query patch descriptors $\mathcal{D}(q)$ for a given query region $q$ are used in the following for our voting strategy.

\textbf{Voting based on local matches.} 
As described previously, we encode each dataset image $I \in \mathcal{I}$ with a set of local patch descriptors $\mathcal{D}(I)$, where we denote the set of all descriptors in the dataset by $\mathcal{D}$.
For a given query rectangle $q$, we also extract a set of local patch descriptor $\mathcal{D}(q)$.
In the following, we do not distinguish between the local patches itself and their feature descriptors, but it should always be clear what is meant from the context.
Our voting consist of the following two main steps.

In the first step, we determine the k-nearest neighbours ${N}\!{N}_k(f,\mathcal{D})$ for each local query patch $f \in \mathcal{D}(q)$ using the search index with $L_2$ distances, where we denote a local query patch with one of its k-nearest neighbours as a local match.
Based on the $L_2$ distances, we define a local matching score via
\begin{equation}
\label{eq:sim_localmatch}
    s_f(g) = \exp\big( -  \| g - f \|_2^2 \, / \, \| \hat g -  f \|_2^2 \big),
\end{equation}
where $\hat g \in {N}\!{N}_k(f,\mathcal{D})$ with a fixed rank and provides a reference distance. 
For the image ranking, we utilize a majority based voting, where we determine the most promising images. To do this, for a given image $I$, we look at its local matches and sum their local matching scores, 
where we consider at most one hit for each local query feature.
We select the images with the highest scores and restrict the following voting based on local matches on this subset. Due to this pre-selection, the computational costs of the following steps are independent of the dataset size. However, in contrast to other approaches \cite{seguin2018replica}, the pre-selection is conducted on a part, not image level, and hence we do not miss similar small regions in the final search results. 

In the second step, we apply our voting scheme on the local matches of the most promising images to predict well-localized retrieval bounding boxes. For the voting, we assume persistent aspect ratios and neglect object rotations to reduce the voting space and accelerate the search. Let us consider a local match $(f,g)$, i.e. $g \in {N}\!{N}_k(f,\mathcal{D}) \cap \mathcal{D}(I)$ for a local query patch $f \in \mathcal{D}(q)$, then this match is voting for a specific location and scale of a rectangle $r$ in image $I$. If we denote $\mathbf{v}(f) = \mathbf{c}_q - \mathbf{c}_f$ as the vector from the center of the local query patch $\mathbf{c}_f$ to the query rectangle $\mathbf{c}_q$ and $d_f$, $d_g$ and $d_q$ are the diagonal lengths of $f$, $g$ and $q$, respectively.
Then this match votes for a rectangle with center $\mathbf{c}_r = \mathbf{c}_g + \mathbf{v}(f) \cdot d_g / d_f $ and diagonal $d_d = d_q \cdot d_g / d_f$.
We aggregate these votes and create a voting map in which each point votes for the center of a box at the corresponding position similar to \cite{shen2012object}.
To keep the voting map compact, we quantize the image space so that the voting map is much smaller than the actual image. 
Here, we reduce quantization errors by voting for a $5 \times 5$ window with a Gaussian kernel for each local match.
For the voting score, we use the local matching similarity defined in Equ. \ref{eq:sim_localmatch}.
To find the correct diagonal lengths for the retrievals, we do not increase the voting space by an additional dimension but average the diagonals of all votes pointing to the same center. This keeps the voting space compact and accelerates the search.  
For determining the center and diagonal of the best retrieval box in the image $I$, we take the position of the maximum in the voting map and the corresponding averaged diagonal. 
This voting approach allows object retrieval and localization at the same time without an exhaustive sliding-window search as a post-processing step, like in \cite{seguin2017tracking,shen2019discovering}.

\textbf{Local query expansion and re-voting.}
After finding our first retrievals across the dataset, we improve our search results using local query expansion and re-voting. 
Concerning the local query expansion, we consider the first ten nearest neighbours in different images for each local query patch and fuse their feature descriptors by taking their mean and $L_2$ normalization over the feature channels. 
By this, we obtain new and more generalized local patch descriptors, containing more diverse information by including multiple instances. The aggregated patches no longer have the same coordinates in the query image because of the combination of local patch descriptors and possible shifts in local matches. Therefore, we also update the voting vectors based on the new patch representations. We do this by determining their nearest neighbours in the query image and measuring the voting vector to the query bounding box center. The generalized local patch queries and updated voting vectors have the same structure as in the first voting stage, and we can apply the previously described voting procedure again, which leads to better search results.


\subsection*{Implementation details}
Concerning the multi-style feature fusion,
we use VGG16 with batch normalization \cite{simonyan2014very}, pre-trained on ImageNet \cite{deng2009imagenet}, and truncated after the fourth layers' RELU as backbone architecture.
We rescale images to $640$ pixels concerning the smallest image side and pad them by $20$ pixels on each side. We add a max-pooling layer and hence obtain a ratio of $16$ between image and feature space. 
For the style transfer model, we use the pre-trained model of Li et al. \cite{li2018learning}, which was trained on the MS-COCO \cite{lin2014microsoft} dataset. We also experimented with training on each target dataset separately, but this did not improve the performance and heavily increased the initialization time.
We generate the stylized images during the offline stage so they only need to be loaded for the feature extraction. 

Concerning the iterative voting, our proposal algorithm uses six different scales with a scaling factor of $2^{-1/2}$. The patch size ranges from $1/12$ up to $1/2$, where we use a stride of $1/50$ regarding the largest image side.
For the selection of discriminative patches, we set $k_e=200$. 
To keep the search index compact, we linearly decrease the number of proposals per image with increasing dataset size. First, we extract 4000 proposals and reduce the number by 375 for each additional 20k images, starting with a dataset size of 20k images. 
The OPQ algorithm utilizes 96 sub-quantizers with 8 bits allocated for each sub-quantizer. For the IVF algorithm, we generate 1024 clusters and use 30 for the nearest neighbor search. 

\section*{Experiments}
In this section, we present comparative evaluations on challenging benchmark datasets and diagnostic experiments. 

\subsection*{Datasets}
To show the efficiency and generalization capability of our system, we evaluate on five different benchmark datasets.

\textbf{Brueghel.} 
Our main evaluation is based on a collection of Brueghel paintings \cite{brueghel} with annotations from Shen et al. \cite{shen2019discovering}. To the best of our knowledge, no other dataset for instance retrieval and localization in the arts with annotations is available at the moment. The dataset consists of 1,587 paintings, including a variety of different techniques, materials, and depicted scenes. It includes ten annotated motifs with 11 up to 57 instances of each motif, which results in 273 annotations overall. We follow the evaluation protocol of \cite{shen2019discovering}, and count retrievals as correct, if the intersection over union (IoU) of predicted with ground-truth bounding boxes is larger than 0.3. For each query, we compute the Average Precision (AP), average these values per class, and report the class level mean Average Precision (mAP).

\textbf{Brueghel5K and Brueghel101K.}
We are particularly interested in the large-scale scenario, where the algorithm has to deal with an extensive and inhomogenous image collection. This is a much more common use case for art historians. For this purpose, we introduce the Brueghel5K and Brueghel101K dataset, where we extend the previously described dataset with an additional 3,500 and 100,000 randomly selected images from the Wikiart dataset \cite{wikiart} as distractors, respectively. To avoid false negatives, we used the annotations from Wikiart and excluded all Brueghel paintings from the selection. The evaluation uses the same annotations and evaluation protocol as the Brueghel dataset.

\textbf{LTLL.} 
We also evaluate our algorithm on the Large Time Lags Location (LTLL) dataset, which was collected by Fernando et al. \cite{fernando2015location}.
It consists of historical and current photos of 25 cities and towns spanning over a range of more than 150 years. 
The main goal is to recognize the location of an old image using annotated modern photographs, where the old and new images can be considered as belonging to two different domains.
In total, the dataset contains 225 historical and 275 modern images.
Since our retrieval system assumes that users mark image regions he is interested in, we provide and utilize additional query bounding boxes for our and all baseline models.
Analogously to the evaluation protocol \cite{fernando2015location} we report the accuracy of the first retrieval.

\textbf{Oxford5K.} 
We also evaluate our approach on the Oxford5K datasets, which was collected by Philbin et al. \cite{philbin2007object}. It consists of 5,062 photos with 11 different landmarks from Oxford and five different query regions for each location. The occurrence of each landmark ranges from 7 up to 220. We follow the evaluation protocol of \cite{philbin2007object} and compute the Average Precision (AP) for each query, average them per landmark and report the mean Average Precision (mAP).

\setlength{\tabcolsep}{4pt}
\begin{table}[tb]
\centering
\caption{
Retrieval performance comparison of our multi-style feature fusion (Ours) and features generated from VGG16, which are either pre-trained on ImageNet \cite{deng2009imagenet} (ImageNet pre-training) or additionally fine-tuned using \cite{shen2019discovering} (Artminer)
}
{
\small
\label{table:ablation-msfeature}
\begin{tabular}{l|c|c|c|c|c}
\hline
Features & Brueghel & Brueghel5K & Brueghel101K & LTLL & Oxford5K \\\hline\hline 
ImageNet pre-training                      &         79.1  &         76.7  &         67.3  &         88.1  &         87.9 \\
Artminer \cite{shen2019discovering}        &         80.6  &         37.9  &         34.5  &         89.0  &         79.4 \\ 
Ours                                       & \textbf{85.7} & \textbf{84.1} & \textbf{76.9} & \textbf{90.9} & \textbf{89.8} \\
\hline
\end{tabular}
}
\end{table}
\setlength{\tabcolsep}{1.4pt}

\subsection*{Effect of multi-style feature fusion}
To validate our multi-style feature fusion, we compare the performance of our algorithm with different feature representations. As baselines, we use VGG16 features truncated after the conv-4 layer, which are either pre-trained on ImageNet \cite{deng2009imagenet} (ImageNet pre-training) or additionally fined-tuned with the self-supervised approach of Shen et al. \cite{shen2019discovering} (Artminer). 

From Tab. \ref{table:ablation-msfeature} it can be seen that our approach improves the results on all benchmarks. The improvement compared to pre-trained features is especially high for the art datasets since there is a particular large domain gap due to differences in styles between queries and targets. 
Even for datasets without any domain shift, like Oxford5K, our algorithm improves the search results due to the aggregation of diverse feature representations. 
We also achieve significantly better results compared to the Artminer \cite{shen2019discovering} fine-tuned variants. Their method has particular problems for image collections containing many distractors as well as on the Oxford5K dataset.

\setlength{\tabcolsep}{4pt}
\begin{table}[tb]
\centering
\caption{
Ablation study of our voting approach.
We measure the performance for searching only with the selected query region (wo/voting), and for restricting ourselves to the first round of voting (wo/it.voting) and our full system. 
We also report the performance on the Brueghel dataset for different IoU thresholds
}
\label{table:ablation-voting}
{
\small
\begin{tabular}{l|ccc|c|c}
\hline
\multirow{2}{*}{Methods} & 
\multicolumn{3}{c|}{Brueghel} & \multirow{2}{*}{LTLL} & \multirow{2}{*}{Oxford5K}
\\\cline{2-4}& IoU@$0.3$ & IoU@$0.5$ & IoU@$0.7$ & & \\
\hline
\hline
Ours wo/voting      &          72.3 &          48.5  &          5.9 &          73.2  &         72.7 \\
Ours wo/it.voting   &          74.1 &          54.4  &         19.1 &          90.4  &         87.8\\
Ours                & \textbf{85.7} &  \textbf{63.3} & \textbf{21.7} & \textbf{90.9} & \textbf{89.8}\\
\hline
\end{tabular}
}
\end{table}
\setlength{\tabcolsep}{1.4pt}

\subsection*{Effect of iterative voting}
We analyze the impact of our voting procedure on the visual search. For this purpose, we measure the performance of our method for searching only with the selected query region (wo/voting), restricting ourselves to the first round of voting (wo/it.voting) and our full system. To better understand the impact on the localization of retrievals, we also report the performance for different IoU thresholds on the Brueghel dataset.  

The results are summarized in Tab. \ref{table:ablation-voting}.  We see that the first round of voting especially improves the results on the LTLL and the Oxford5K dataset, the performance gain on the Brueghel dataset for the IoU of 0.3 is smaller. 
The reason is that the query regions for the Brueghel dataset are, on average, much smaller. Therefore, there are fewer voting regions, and the effect of voting becreases. However, the results for higher IoU thresholds show that the localization is significantly improving. The second round of voting with local query expansion improves the results on all datasets further. 
This has a particularly strong influence on the retrieval results for the Brueghel dataset.

\subsection*{Computational cost}
We also investigate the search speed of our Python implementation. 
All measurements are conducted on the same machine with 3 GPUs (Nvidia Quadro P5000). In Tab. \ref{table:computational-cost}, we summarize the results and compare them with \cite{shen2019discovering}, where we assumed a perfect implementation of \cite{shen2019discovering} on multiple GPUs by dividing their search times by a factor of 3. 
Besides the search speed, we also report the index size after storing to disk. 
\setlength{\tabcolsep}{4pt}
\begin{table}[tb]
\centering
\caption{
Comparison of the retrieval time of our method (Ours) and \cite{shen2019discovering} (Artminer) for different dataset sizes, where we also report the size of our search index (last row)
}
{
\small
\label{table:computational-cost}
\begin{tabular}{l|c|c|c|c|c|c}
\hline
Method & 5K & 20K & 40K & 60K & 80K & 100K \\\hline\hline 
Artminer \cite{shen2019discovering} & 12.9 min & 50.4 min & 1.7 h & 2.8 h &  3.8 h & 4.6 h \\
Ours                                 & 8.5 s& 9.0 s& 9.3 s & 9.7 s &  10.1 s & 10.5 s \\
\hline
\hline
Ours                                 & 1.7 GB & 6.8 GB & 13.7 GB & 20.6 GB &  27.4 GB & 34.2 GB \\
\hline
\end{tabular}
}
\end{table}
\setlength{\tabcolsep}{1.4pt}
It shows that our method is much faster, and its speed depends only moderately on the number of images. The index size mainly determines the size of the image collection that can be searched.
Its size primarily depends on the number of proposals extracted per image, which is an important factor in finding small regions. Since the required retrieval accuracy for small regions,  available hardware, and dataset size vary from application to application, the number of proposals should be adjusted according to the actual use case. 

\setlength{\tabcolsep}{4pt}
\begin{table}[tb]
\centering
\caption{
Retrieval results of our method and state-of-the-art methods on the Brueghel \cite{brueghel,shen2019discovering}, Brueghel5K, Brueghel101K, LTLL \cite{fernando2015location} and Oxford5K \cite{philbin2007object} dataset. We also report the underlying network architecture (Net), and what dimension the underlying features have (Dim)
}
\label{table:performance}
\centering
{
\small
\begin{tabular}{l|c|c|c|c|c|c|c}
\hline
\multirow{2}{*}{Methods} & 
\multirow{2}{*}{Net} &
\multirow{2}{*}{Dim} &
\multicolumn{3}{c|}{Brueghel} & 
\multirow{2}{*}{LTLL} & \multirow{2}{*}{Oxford}
\\\cline{4-6} & & & \cite{brueghel,shen2019discovering} & 5K & 101K & & \\
\hline
\hline
ImageNet, image level & VGG16 & 512 & 24.0         & 22.5 & 17.7  & 47.8           & 25.6  \\
Radenovi\'c et al. \cite{radenovic2018fine}, wo/ft  & VGG16           & 512 & 15.5          & 12.7 & 5.9 & 59.3           & 53.4 \\
 Radenovi\'c et al. \cite{radenovic2018fine}            & VGG16 & 512 & 15.8          & 12.8 & 5.7 & 76.1           & 87.8 \\
Artminer \cite{shen2019discovering}, wo/ft   & ResNet18  & 256  & 58.1  & 56.0  & 50.2  & 78.9  & 84.9  \\
Artminer \cite{shen2019discovering}                 & ResNet18  & 256  & 76.4  & 46.5  & 37.4  & 88.5  & 85.7  \\
Artminer \cite{shen2019discovering}, wo/ft   & VGG16   & 512  & 54.4  & 50.5  & 44.1  & 81.8  & 85.0  \\
Artminer \cite{shen2019discovering}              & VGG16   & 512  & 79.9  & 39.5  & 36.4  & 88.9  & 81.5   \\\hline
Ours   & VGG16 & 96  &         85.7  &         83.9  & \textbf{76.9}   &         90.9          & \textbf{89.9} \\
Ours   & VGG16 & 128 &         87.2  &         85.6  & ----            &         90.9          &         89.8 \\
Ours   & VGG16 & 256 & \textbf{88.1} & \textbf{86.7} & ----            & \textbf{91.3}         &         89.8 \\
\hline
\end{tabular}
}
\end{table}
\setlength{\tabcolsep}{1.4pt}

\begin{figure}[tb]
\centering
\begin{minipage}[t]{0.495\textwidth}
\centering
\begin{places}{1.0\textwidth}{6}
\place{\scriptsize Ours}{
  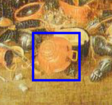,
  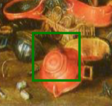,
  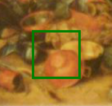,
  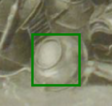,
  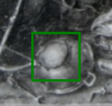,
  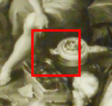
}
\place{\scriptsize Artminer}{
  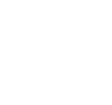,
  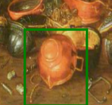,
  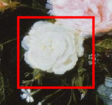,
  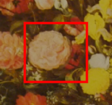,
  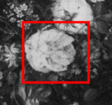,
  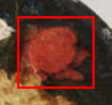
}
\place{\scriptsize Ours}{
   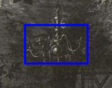,
  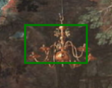,
  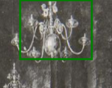,
  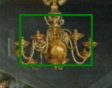,
  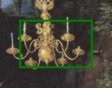,
  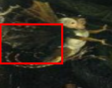
}
\place{\scriptsize Artminer}{
   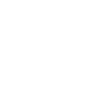,
  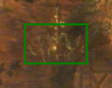,
  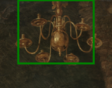,
  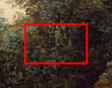,
  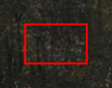,
  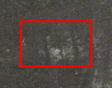
}
\end{places}
\end{minipage}
\begin{minipage}[t]{0.495\textwidth}
\centering
\begin{places}{1.0\textwidth}{6}
\place{\scriptsize Ours}{
   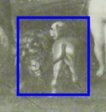,
  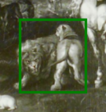,
  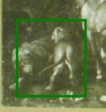,
  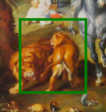,
  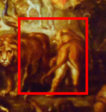,
  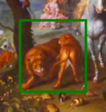
}
\place{\scriptsize Artminer}{
   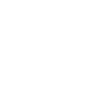,
  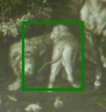,
  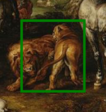,
  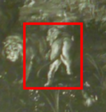,
  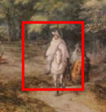,
  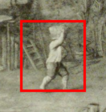
}
\place{\scriptsize Ours}{
   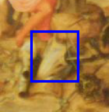,
  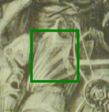,
  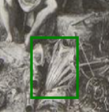,
  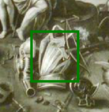,
  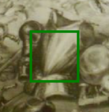,
  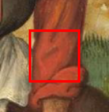
}
\place{\scriptsize Artminer}{
   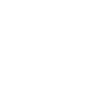,
  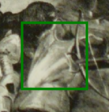,
  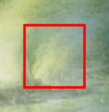,
  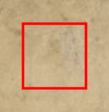,
  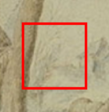,
  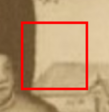
}
\end{places}
\end{minipage}
\\
\begin{minipage}[t]{0.495\textwidth}
\hspace{1.1cm}
\includegraphics[width=0.7\textwidth]{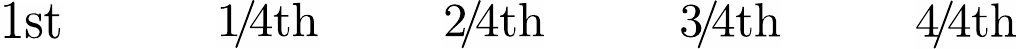}
\end{minipage}
\begin{minipage}[t]{0.495\textwidth}
\hspace{1.1cm}
\includegraphics[width=0.7\textwidth]{results/brueghel/numbers/numbering.pdf}
\end{minipage}
\caption{Qualitative comparison.
Retrieval examples of our approach (Ours) and \cite{shen2019discovering} (Artminer) on the Brueghel dataset. 
We show queries on the left in blue and its nearest-neighbor, as well as four additional retrievals with an equidistant distance given by the number of ground truth annotations for the query divided by four. 
We draw green bounding boxes if the intersection over union is larger than 0.3 and red otherwise.
}
\label{fig:qualitative_comparison}
\end{figure}

\subsection*{Retrieval performance on benchmarks}
In the following, we give quantitative and qualitative results on the previously introduced benchmark datasets.

\textbf{Quantitative evaluation.}
We compare our results with max-pooled pre-trained features on image level 
(ImageNet, image level) as well as the state-of-the-art results of Shen et al. \cite{shen2019discovering} (Artminer) and Radenovi\'c et al. \cite{radenovic2018fine}
on Brueghel, LTLL, and Oxford5K. For a fair comparison, we select the same backbone architecture for all methods and report also the original numbers from \cite{shen2019discovering} with their fine-tuned ResNet18 model. Our approach utilizes marked query regions within the image. This is not the case for the discovery mode of \cite{shen2019discovering}. According to their publication and our experiments, they obtain the best results using full images as query, which allows their algorithm to utilize more context. The reported numbers refer to their discovery mode. 

We summarize all results in Tab \ref{table:performance}. 
We outperform all methods on all benchmark datasets without fine-tuning on the retrieval task and with a much smaller feature dimension. 
The results on the Brueghel datasets with additional distractors show that the self-supervised method of Shen et al. \cite{shen2019discovering} is not stable against images without corresponding regions in the dataset. The main reason is that, the probability of selecting regions without correspondences is very high, which results in few and potentially spurious matches for training. This effect can already be seen for Brueghel5K, where their fine-tuned network leads to worse retrieval results compared to their initial model. In contrast, our method is much more robust against such distractors. Furthermore, it can be seen that we even outperform \cite{radenovic2018fine} on the Oxford5K dataset, although, their approach is explicitly designed for geo-localization by fine-tuning on an extensive image collection of various landmarks using ResNet101 \cite{he2016deep}. However, since their model is optimized for this task, their search results on art collections are rather weak.

\begin{figure}[t]
\centering
\begin{minipage}[t]{0.495\textwidth}
\centering
\begin{places}{1.0\textwidth}{6}
\place{\scriptsize Full}{
   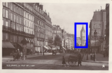,
  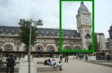,
  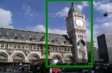,
  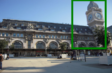,
  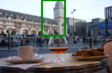,
  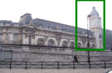
}
\vspace{0.15cm}
\place{\scriptsize Zoom}{
   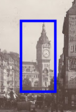,
  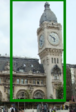,
  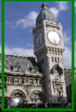,
  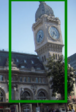,
  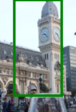,
  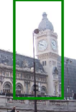
}
\place{\scriptsize Full}{
   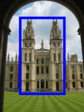,
  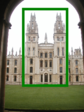,
  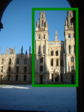,
  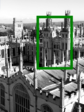,
  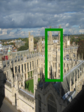,
  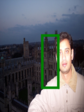
}
\place{\scriptsize Zoom}{
   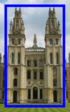,
  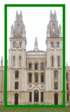,
  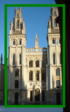,
  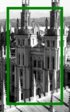,
  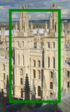,
  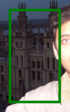
}
\end{places}
\end{minipage}
\begin{minipage}[t]{0.495\textwidth}
\centering
\begin{places}{1.0\textwidth}{6}
\place{\scriptsize Full}{
   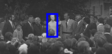,
  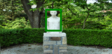,
  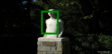,
  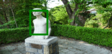,
  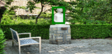,
  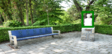
}
\vspace{0.15cm}
\place{\scriptsize Zoom}{
   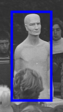,
  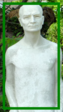,
  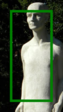,
  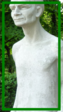,
  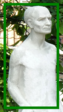,
  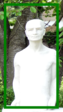
}
\place{\scriptsize Full}{
   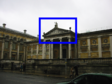,
  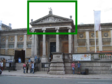,
  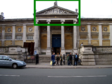,
  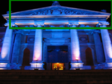,
  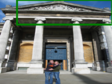,
  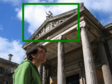
}
\place{\scriptsize Zoom}{
   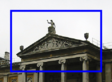,
  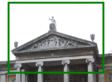,
  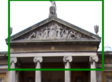,
  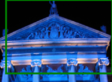,
  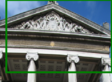,
  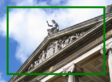
}
\end{places}
\end{minipage}
\\
\begin{minipage}[t]{0.495\textwidth}
\hspace{1.2cm}
\includegraphics[width=0.68\textwidth]{results/brueghel/numbers/numbering.pdf}
\end{minipage}
\begin{minipage}[t]{0.495\textwidth}
\hspace{1.2cm}
\includegraphics[width=0.68\textwidth]{results/brueghel/numbers/numbering.pdf}
\end{minipage}
\caption{
Retrieval examples.
We show examples in rows 1-2 and 3-4 for the LTLL and Oxford5K dataset. The first row shows full images (Full) and the second zoomed-in versions (Zoom).  
The queries are visualized in blue on the left and its nearest-neighbor, as well as four additional retrievals with an equidistant distance, given by the number of ground truth annotations for the query divided by four on the right. 
}
\label{fig:qualitative_examples}
\end{figure}

\textbf{Qualitative evaluation.}
In Fig. \ref{fig:qualitative_comparison}, we provide a qualitative comparison with the state-of-the-art of Shen et al. \cite{shen2019discovering} (Artminer). It shows that their method can find first retrievals quite well. However, these become significantly worse for higher ranks, where our approach gives much better results. 
In Fig. \ref{fig:qualitative_examples}, we show some qualitative examples for the other datasets. The retrieval results show that our system is capable of finding similar objects despite differences in colour and style. Furthermore, we see that objects can be precisely located despite changes in perspective and partial occlusions, which is also the case for small regions.

\section*{Conclusion}
We have presented a novel search algorithm to find and localize motifs or objects in an extensive art collection. This enables art historians to explore large datasets to find visual relationships. Our algorithm is based on a new multi-style feature fusion, which reduces the domain gap and thus improves instance retrieval across artworks. In contrast to previous methods, we require neither object annotations, image labels, nor time-consuming self-supervised training. The presented iterative voting with recent GPU-accelerated approximate nearest-neighbor search \cite{faiss17} enables us to find and localize even small motifs within an extensive database in a few seconds. We have validated the performance of our model on diverse benchmark datasets, including art collections \cite{brueghel,shen2019discovering} and real photos \cite{philbin2007object}. We have also shown that our method is much more stable against distractors compared to the current state-of-the-art.

\section*{Acknowledgement}
This work has been funded in part by the German Research Foundation (DFG) - project 421703927.

\clearpage
%
%
\bibliographystyle{splncs04}
\bibliography{egbib}
\end{document}